\documentclass[twocolumn]{IEEEtran}

\usepackage{times}
\usepackage{epsfig}
\usepackage{graphicx}
\usepackage{amsmath}
\usepackage{amssymb}
\usepackage{booktabs}
\usepackage{epsfig}
\usepackage{mathrsfs}
\usepackage{multirow}
\usepackage{bm}
\usepackage{bbm}
\usepackage{amsfonts}
\usepackage{supertabular}
\usepackage{array}
\usepackage{gensymb}
\usepackage{threeparttable}
\usepackage{caption}

\usepackage[pagebackref=true,breaklinks=true,letterpaper=true,colorlinks,bookmarks=false]{hyperref}

\usepackage[capitalize]{cleveref}
\crefname{section}{Sec.}{Secs.}
\Crefname{section}{Section}{Sections}
\Crefname{table}{Table}{Tables}
\crefname{table}{Tab.}{Tabs.}

\begin{document}

\newcolumntype{L}[1]{>{\raggedright\arraybackslash}p{#1}}
\newcolumntype{C}[1]{>{\centering\arraybackslash}p{#1}}
\newcolumntype{R}[1]{>{\raggedleft\arraybackslash}p{#1}}

\newcommand{\etal}{\textit{et al}.}

\title{Zero-shot Text-driven Physically Interpretable Face Editing}

\author{Yapeng~Meng,
        Songru~Yang,
        Xu~Hu,
        Rui~Zhao,
        Lincheng~Li,
        Zhenwei~Shi,~\IEEEmembership{Member,~IEEE}
        and~Zhengxia~Zou

\IEEEcompsocitemizethanks{
\IEEEcompsocthanksitem Y. Meng, S. Yang, X. Hu, and Z. Shi are with Image Processing Center, School of Astronautics, Beihang University, Beijing 100191, China, and with Beijing Key Laboratory of Digital Media, Beihang University, Beijing 100191, China, and with State Key Laboratory of Virtual Reality Technology and Systems, Beihang University, Beijing 100191, China, and also with Shanghai Artificial Intelligence Laboratory.
\IEEEcompsocthanksitem R. Zhao, L. Li are with the Netease Fuxi AI Lab, Hangzhou, China.
\IEEEcompsocthanksitem Z. Zou is with the Department of Guidance, Navigation and Control, School of Astronautics, Beihang University, Beijing 100191, China, and also with Shanghai Artificial Intelligence Laboratory.\protect

} % \IEEEcompsocitemizethanks

} % \author

\maketitle
%%%%%%%%% ABSTRACT
\begin{abstract}
This paper proposes a novel and physically interpretable method for face editing based on arbitrary text prompts. Different from previous GAN-inversion-based face editing methods that manipulate the latent space of GANs, or diffusion-based methods that model image manipulation as a reverse diffusion process, we regard the face editing process as imposing vector flow fields on face images, representing the offset of spatial coordinates and color for each image pixel. Under the above-proposed paradigm, we represent the vector flow field in two ways: 1) explicitly represent the flow vectors with rasterized tensors, and 2) implicitly parameterize the flow vectors as continuous, smooth, and resolution-agnostic neural fields, by leveraging the recent advances of implicit neural representations. The flow vectors are iteratively optimized under the guidance of the pre-trained Contrastive Language-Image Pretraining~(CLIP) model by maximizing the correlation between the edited image and the text prompt. We also propose a learning-based one-shot face editing framework, which is fast and adaptable to any text prompt input.  
% We evaluate our method on a rich variety of text prompts and images, including real faces, cartoons, animals, etc. 
Our method can also be flexibly extended to real-time video face editing. Compared with state-of-the-art text-driven face editing methods, our method can generate physically interpretable face editing results with high identity consistency and image quality. Our code will be made publicly available.
\end{abstract}

\begin{figure*}
    \centering{\includegraphics[width=1.0\linewidth]{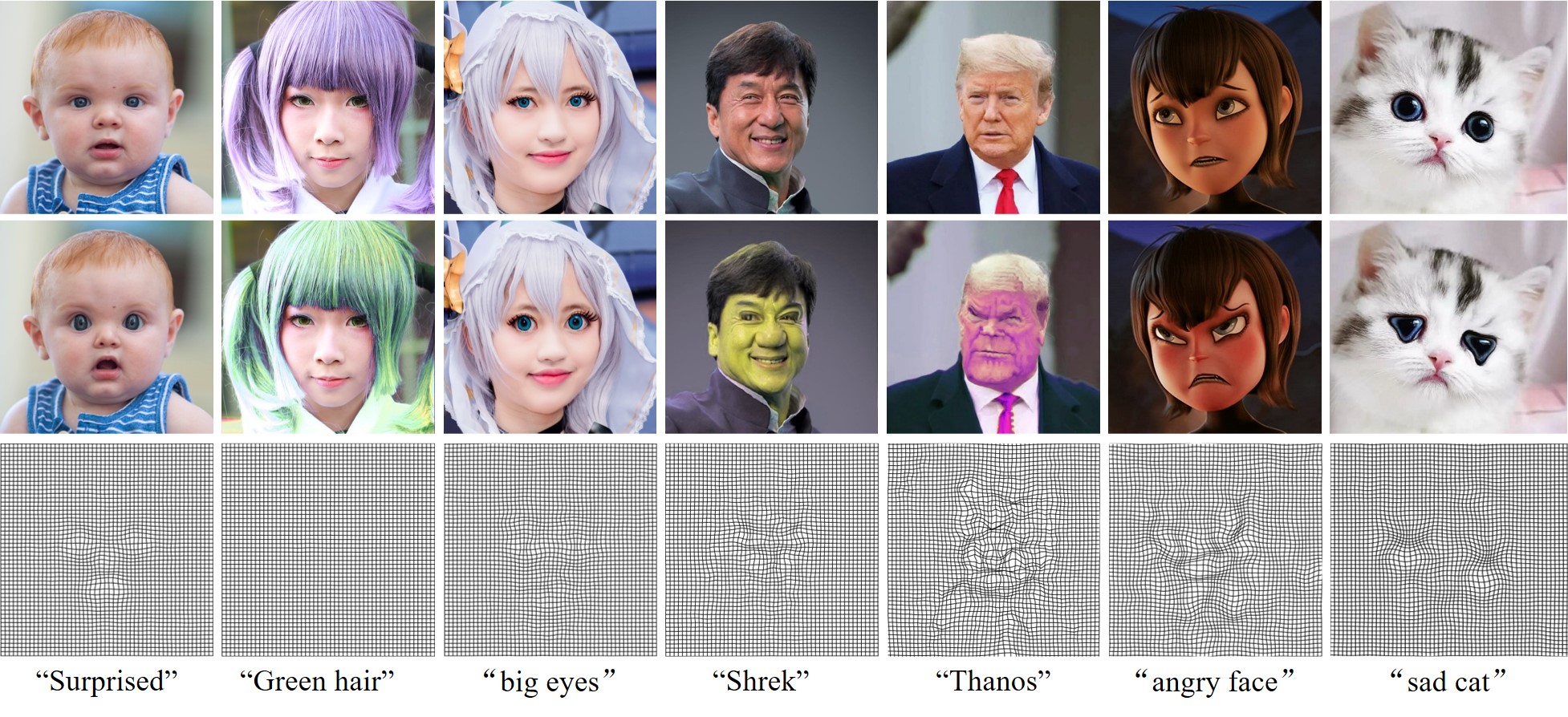}}\\
     \caption{We propose a text-driven physically interpretable face editing method. The face editing process is regarded as imposing vector flow fields on the ordinary image, which represent the spatial and color offset of each pixel. Our method is not only applicable to real human face images, but also to cartoon faces and animal faces. The 1st - 3rd rows show the input images, face editing results (conditioned by the text prompt below), and spatial vector flow fields, respectively.}
    \label{fig:teaser}
\end{figure*}

%%%%%%%%% BODY TEXT
\section{Introduction}

% （对人脸编辑方法、优劣的综述）

\IEEEPARstart{F}{ace} editing is a fundamental research direction in computer vision, which aims at modifying input facial images with the guidance of fixed templates, reference images~\cite{Shi_TPAMI2022}, sketches~\cite{portenier2018faceshop}, semantics~\cite{Shen_2020_CVPR}, or text prompts~\cite{patashnik2021styleclip, Kim_2022_CVPR}. Face editing is a core feature in photo editing software, the film industry, and games. Among them, text-driven face editing, i.e., modifying face images according to input text prompts, is an emerging research topic and has drawn increasing attention recently~\cite{xia2021tedigan, patashnik2021styleclip, Kim_2022_CVPR}. Most of the previous studies in text-driven face editing leverage rich and diverse priors in pre-trained Generative Adversarial Networks (GANs), by editing the latent space of GANs, which are also known as ``GAN-inversion-based methods''. However, previous methods heavily rely on pre-trained large-scale GAN models and per-category training (e.g., ``faces'', ``cars'', ``churches'', etc need to be trained separately on their respective datasets), which limits its performance on generating images out of the pre-trained domain. Recently, diffusion-based method\cite{Kim_2022_CVPR} can also achieve text-driven face editing and have a better generalization ability to unseen domains. However, the face editing process in these approaches also lacks clear physical meaning, causing challenges to identity consistency before and after face editing.

% 在本文中，我们提出一种基于任意提示文本的face editing方法，且warping过程具有physical significance。我们的方法利用CLIP评估生成图像与指定text prompt间的相似度。我们将face editing过程视为图像各个像素的空间偏移确保了整个过程在物理上可解释。

In this paper, we propose a novel face-editing method based on arbitrary text prompts with clear interpretability and physical significance. Different from previous GAN-inversion-based or diffusion-based methods, we regard the face editing process as imposing vector flow fields on the face image, which represent the spatial and color offset of each pixel. 

Our method uses the Contrastive Language-Image Pretraining~(CLIP) model~\cite{radford2021learning} as a guidance network. 
% which is pre-trained on 400 million image and text pairs. 
The guidance model CLIP maps the input text prompt and the edited face image into a unified latent embedding space. The proposed methods can be considered as a text-image correlation maximization process, i.e., by maximizing the similarity between the generated image and the given text prompt within the CLIP embedding space. To generate the flow vectors, we parameterize and optimize them under an end-to-end differentiable framework. We proposed to represent the flow vectors in two approaches. In the first approach, we explicitly represent the flow vectors with rasterized tensors in which the spatial offsets for each pixel are directly recorded. In the second approach, we leverage the recent advances in Implicit Neural Representations (INR) and implicitly parameterize the flow vectors as continuous, smooth, and resolution-agnostic neural fields. We show that both approaches have their own advantages, where the former can represent more detailed deformation while the latter is faster to fit and it has much fewer parameters. Meanwhile, the output of INR is sampled from a continuous function with pixel coordinates as independent variables, which naturally ensures the smoothness of the vector flow field in an image.

In our method, the vector flow fields are generated using iterative optimization or one-shot fast editing. 1) Iterative optimization. Without training any network parameters, with any input face image and text prompt, the rasterized vector flow field is iteratively optimized with the gradient descent under the supervision of the pre-trained CLIP model. The training is performed by maximizing the similarity between the warped face image and the text prompt in the CLIP embedding space. Since the image warping can be performed in a differentiable way, the rasterized vector field can be end-to-end optimized based on the gradients back-propagated all the way from CLIP latent space. 2) One-shot learning for any text prompts. 
% Given a fixed text template or an arbitrary text prompt, we introduce a hyper-network and enforce the network to imitate the behavior of the optimization method by training on paired data (text, vector flow field). Instead of directly generating the rasterized vector field, the hyper-network is trained to predict the weight of the neural field network, which greatly reduces the dimension of the parameter space and the difficulty of learning. 
Given an input face image and an arbitrary text prompt, we introduce a one-shot network to generate a spatial and color editing vector flow field in a single forward step. Instead of using paired data (text, vector flow field) and the supervised method, we use unsupervised learning under the guidance of the pre-trained CLIP model. In the training stage, only simple facial feature descriptions are fed, but in the inference stage, it can be generalized to much more complex descriptions or unseen text prompts.

With the proposed method, we obtained vivid, realistic, and physically meaningful face editing results that conform to the text prompt. Besides, our method can be flexibly extended to cartoon face editing, animal face editing, and video face editing.  Compared with state-of-the-art text-driven face editing methods, our method can generate physically interpretable face editing results with high identity consistency and image quality.

The contribution of our paper is summarized as follows:
\begin{itemize}
\item We propose a novel method for text-driven face editing. Our method is based on differentiable image warping, and as a by-product, our method can also generate physically meaningful pixel-wise vector flow fields.
\item We propose to represent the flow vectors as a continuous neural field based on the implicit neural representation, which naturally guarantees many advantages such as continuity, smoothness, and resolution-animosity.
\item We propose a one-shot method that can generate the vector flow field for arbitrary text prompts in a faster way. It can even generalize to unseen text prompts. 
% Our one-shot model takes full advantage of the consistency of text and image in the latent embedding space from the CLIP encoder.

\end{itemize}

\begin{figure*}
    \centering{\includegraphics[width=1.0\linewidth]{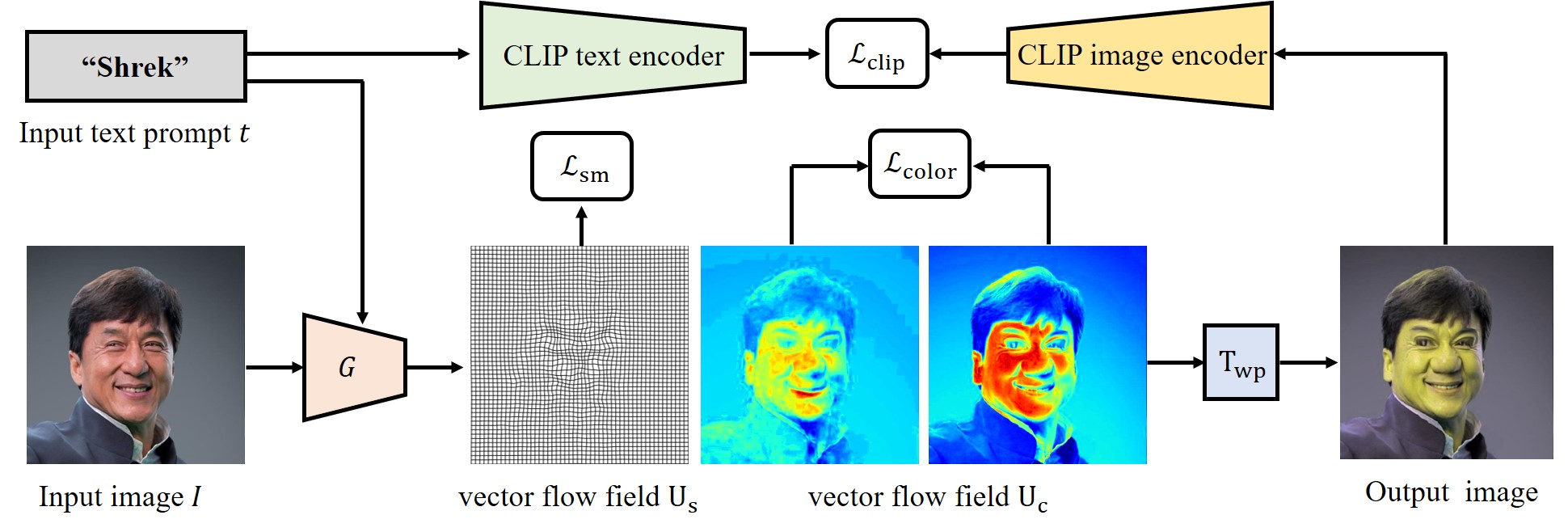}}\\
     \caption{An overview of our proposed method. The image editing processing in our method is regarded as imposing vector flow fields ($U_s$ and $U_c$) on the input image, where $U_s$ is the flow field for spatial coordinate changes and $U_c$ is the field for color changes. A pre-trained CLIP model is employed to ensure the face image after transformation is semantically consistent with the given text prompt.}
    \label{fig:overview}
\end{figure*}

\section{Related Work}
% 由于GAN网络生成 photo-realistic images 的特性，一些研究基于GAN-inversion实现人脸编辑。Shen et al. \cite{Shen_2020_CVPR} conduct a detailed study on how different semantics are encoded in the latent space of GANs and achieve semantic face editing.

{\bf Face Editing.} 
% Face editing used in commercial software mostly uses traditional image processing methods, based on manual liquefaction brushes or moving the face key points by fixed templates. Among them, the manual editing method is difficult to operate for inexperienced users, and the fixed template limits the flexibility of face editing. With the development of deep learning and GAN~\cite{goodfellow2014generative}, researchers have studied more flexible face editing methods. For example, face editing based on the conditional input such as sketch~\cite{portenier2018faceshop}, semantic~\cite{Shen_2020_CVPR}, reference image~\cite{Shi_TPAMI2022}, text prompt~\cite{patashnik2021styleclip}, and etc. 
Leveraging the advantages of GAN in high-fidelity image generation, e.g., StyleGAN~\cite{karras2019style}, most recent studies implement face editing based on GAN inversion. Shen \etal~\cite{Shen_2020_CVPR} studies how different semantics are encoded in the latent space of GANs and achieve semantic face editing. Instead of discovering semantically meaningful latent manipulations by human examination, Patashnik \etal propose StyleCLIP~\cite{patashnik2021styleclip} and implement text-driven manipulation of StyleGAN imagery by utilizing a CLIP-based loss to modify the input latent vector. However, due to the limited GAN inversion performance and pre-trained domain, these methods are difficult to preserve the original image details and also fail to generate images in unseen domains. Diffusion-based models have better performance on unseen domains, Kim \etal~\cite{Kim_2022_CVPR} propose DiffusionClip, achieving text-driven image manipulation by fine-tuning a pre-trained diffusion model under the guidance of CLIP-based loss, but having difficulty in keeping human identity consistency.
% Wang \etal~\cite{wang2022rewriting} propose to synthesize things that go far beyond the data distribution and everyday experience, where they enable a user to ``warp'' a given model by editing only a handful of original model outputs with desired geometric changes. 
Different from the above methods dealing with face editing as a black box process, we investigate generating physical explainable flow vectors that enable ``wiz-ee-wig'' face editing.

{\bf Implicit Neural Representation.} Recently, the use of neural networks to approximate continuous space and temporal functions, i.e., Implicit Neural Representations (INR)~\cite{mildenhall2020nerf, INRChen_CVPR21, sitzmann2020implicit}, has become an emerging research topic in machine learning and computer vision. In INR, an object is usually represented as a multilayer perceptron (MLP) that maps coordinates to signal. NERF~\cite{mildenhall2020nerf} learns an implicit representation for a specific scene using multiple image views and performs novel view synthesis. INR has been widely used in many applications such as novel view synthesis~\cite{mildenhall2020nerf}, image super-resolution~\cite{INRChen_CVPR21}, and image synthesis~\cite{skorokhodov2021aligning}. Objects represented by the INR network are resolution-independent, which can generate arbitrary resolution rasterized output.
% Also, there are some approaches~\cite{nair2010rectified} focusing on improving the expression ability of INR and speeding up the training. For example, Sitzmann et al.~\cite{sitzmann2020implicit} leverage periodic activation functions to replace the common ReLU~\cite{relu} activation function in MLP, which makes the network capable of modeling signals with fine detail. Tancik \etal~\cite{tancik2020fourfeat} explore positional encoding methods for input coordinates and use Fourier Feature Mapping which enables an MLP to learn high-frequency functions in low-dimensional problem domains. 
In this paper, we also take advantage of the INR and explore the latest INR techniques in face editing.

{\bf Contrastive Language-Image Pretraining (CLIP).} CLIP~\cite{radford2021learning} is a recently proposed open-source large-scale multimodal pre-train model by OpenAI for image classification. CLIP is trained on 400 million image and text pairs, connecting them in a unified latent embedding space. Its power of aligning vision and language has been exploited in many recent studies~\cite{patashnik2021styleclip, perez2021generating, wang2022clip, Kim_2022_CVPR}. Similar to our method, some recent methods in generative modeling integrate CLIP for text prompt guidance. For example, CLIP-NeRF~\cite{wang2022clip} uses the CLIP encoder to extract feature embedding for editing neural radiance fields in a view-consistent way. Dream Fields~\cite{jain2022DreamFields} generates open-set 3D models from natural language prompts under the guidance of the CLIP model. 
% DiffusionCLIP~\cite{Kim_2022_CVPR} combines a diffusion model~\cite{song2019diffusion} with CLIP to conduct a text-driven image manipulation.

%------------------------------------------------------------------------
\section{Methodology}

\subsection{Overview}
Fig.~\ref{fig:overview} shows an overview of our method. Given an image $I$ and a text prompt $t$, we generate semantically consistent vector flow fields $U_s$ and $U_c$ to warp the input image, where $U_s$ is the flow field for spatial coordinate changes and $U_c$ is the flow field for color changes. The overall process is defined as follows:
\begin{equation}\label{eq:overview_equation}
\begin{split}
    & I^\prime = T_{wp}(I, U_s, U_c), \\
    & U_s, U_c = G(I, t),
\end{split}
\end{equation}
where $T_{wp}$ is the warping operation, detailed in Sec.~\ref{Sec: warp}, $G$ is the flow field generation process, detailed in Sec.~\ref{section:Iterative optimization of INR} and Sec.~\ref{section:One-shot learning for each prompt}. 
% , consists of spatial offset and color changing. 

A pre-trained CLIP~\cite{radford2021learning} model is employed to ensure the vector flow field is semantically consistent with the given text prompt. We aim at searching for an optimal field $(U_s^*, U_c^*)$ so that the image after warping has the maximum similarity with the text prompt in the CLIP embedding space:
\begin{equation}\label{eq:opt}
\begin{split}
U_s^*, U_c^* &= \arg \min_{U_s, U_c} \mathcal{L}(I, t, U_s, U_c), \\
\mathcal{L}_{clip}(I, t, U_s, U_c) &= -s(E(T_{wp}(I, U_s, U_c)), E(t)),\\
\mathcal{L}(I, t, U_s, U_c) &= \lambda_{clip}\mathcal{L}_{clip} + \lambda_{sm}\mathcal{L}_{sm}(U_s) \\
                            &+ \lambda_{color}\mathcal{L}_{color}(U_c) \\
                            &+ \lambda_{id}\mathcal{L}_{id}(I, T_{wp}(I, U_s, U_c)),
\end{split}
\end{equation}
where $E$ is a pre-trained CLIP image/text encoder, $s(\cdot)$ is a similarity measurement function in CLIP embedding space. Smoothness loss $\mathcal{L}_{sm}$ is used to control the continuity of the vector flow field. Color loss $\mathcal{L}_{color}$ is used to limit colors to changing not too much. Identity loss $\mathcal{L}_{id}$ is used to keep face identity consistency.

Two modes, ``iterative optimization'' and ``one-shot learning'', are both explored to generate the vector flow field.
In the following, we first introduce the vector flow field and then give a detailed introduction of the two generation modes.
  
% The INR network learns the parametric representation of the vector flow field to enable continuous, smooth, and resolution-independent warping. 
% We train the proposed model in two ways: iterative optimization and one-shot learning respectively.

\begin{figure*}
    \centering{\includegraphics[width=1.0\linewidth]{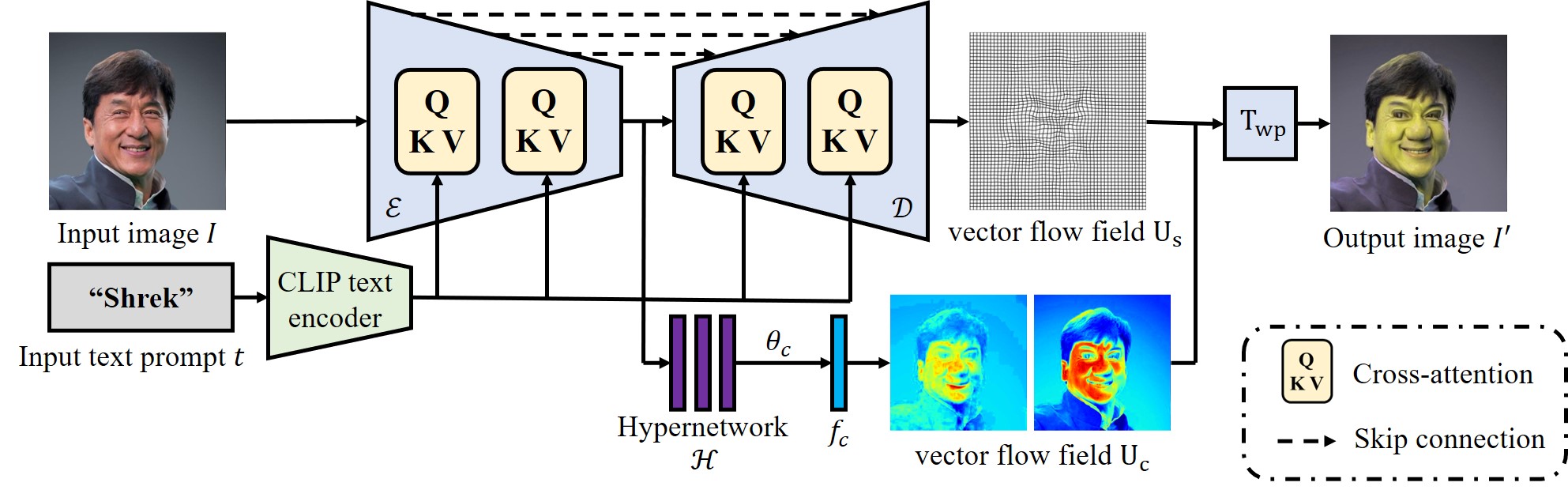}}\\
     \caption{One-shot learning mode of our method. It's a UNet-based encoder-decoder network. Text prompt $t$ is encoded by the CLIP text encoder and mapped to the intermediate layers of the UNet by the cross-attention mechanism. A hyper-network $\mathcal{H}$ is used to predict the parameters $\theta_c$ of the color flow field $f_c$.
     % The image editing processing in our method is regarded as imposing vector flow fields ($U_s$ and $U_c$) on the input image, where $U_s$ is the flow field for spatial coordinate changes and $U_c$ is the field for color changes. A pre-trained CLIP model is employed to ensure the face image after transformation is semantically consistent with the given text prompt.
     }
    \label{fig:method_oneshot}
\end{figure*}

\subsection{Warping image by the vector flow field}\label{Sec: warp}

Our method edits face images based on the vector flow field. $T_{wp}$ includes image coordinate offsets and color changes. The spatial flow field $U_s$ represents the coordinate offsets of each pixel at $(x,y)$ location:
\begin{equation}
    I_{x,y}^\prime = I_{x+\delta x, y+\delta y},
\end{equation}
where $\delta x, \delta y = U_s(x,y)$.
For pixel colors, we also define a similar ``flow field'' $U_c$ that represents the changes of the color value $c$ at location $(x,y)$:
\begin{equation}
    c_{x,y}^\prime = c_{x,y} + \delta c_{x,y},
\end{equation}
where $\delta c_{x,y}=U_c(x,y)$. Since Implicit Neural Representations (INR) can naturally represent continuous, smooth, and resolution-independent field signals, we parameterize the fields $U_s$ and $U_c$ with two neural networks $f_s$ and $f_c$. Given any pixel coordinate $(x,y)$, we define $f_s(x,y|\theta_s)\approx U_s(x,y)$ and $f_c(x,y|\theta_c)\approx U_c(x,y)$, where $\theta_s$ and $\theta_c$ are trainable parameters of the neural networks. Inspired by the NeRF~\cite{mildenhall2020nerf}, we add high dimensional position encoding to enable the INR network to learn higher frequency details. This helps the INR network converge rapidly and efficiently with fewer parameters. With the INR representation, the problem (\ref{eq:opt}) can be reformulated as seeking the best neural network parameters:
\begin{equation}\label{eq:opt_inr}
\begin{split}
\theta_s^*, \theta_c^* &= \arg \min_{\theta_s, \theta_c} \mathcal{L}(I, t, f_s, f_c), \\
\mathcal{L}_{clip}(I, t, f_s, f_c) &= -s(E(T_{wp}(I, f_s(\theta_s), f_c(\theta_c))), E(t)), \\
\mathcal{L}(I, t, f_s, f_c) &=\lambda_{clip}\mathcal{L}_{clip} + \lambda_{sm}\mathcal{L}_{sm}(f_s(\theta_s)) \\
                            &+ \lambda_{reg}\mathcal{L}_{reg}(f_s(\theta_s)) +\lambda_{color}\mathcal{L}_{color}(f_c(\theta_c))\\
                            &+ \lambda_{id}\mathcal{L}_{id}(I,T_{wp}(I, f_s(\theta_s), f_c(\theta_c))),
\end{split}
\end{equation}
here we add a regular loss term $\mathcal{L}_{reg}$ to limit the overall complexity of the vector flow field.

% \begin{figure}[t]
%     \centering{\includegraphics[width=1.0\linewidth]{images/method-oneshot.jpg}}\\
%     \caption{One-shot learning mode of our method. We train a hyper-network to predict the parameters of the INR network.
%     % vector flow field to reflect differences between the encoded results of image and text prompt in CLIP Latent Embedding Space. 
%     (a) In the training stage, the hyper-network takes in not only the input image but also the warped image. (b) In the inference stage, the hyper-network takes in the input image and text prompt. (c) Detailed architecture of our hyper-network and INR network.}
%     \label{fig:method_oneshot}
% \end{figure}

\subsection{Iterative optimization mode}\label{section:Iterative optimization of INR}

Fig.~\ref{fig:overview} shows the process of our iterative vector field optimization. The semantic consistency between generated vector flow field and the given text prompt is constrained by a pre-trained CLIP model. We define the similarity measurement in Eq.~(\ref{eq:opt_inr}) as the cosine distance between the CLIP embeddings of the input warped image and text prompt. Gradient descent is used to update the parameters of the INR network:
\begin{equation}\label{eq:bp}
\begin{split}
    \theta_s &\leftarrow \theta_s + \partial \mathcal{L}(I, t, f_s, f_c) / \partial \theta_s\\
    \theta_c &\leftarrow \theta_c + \partial \mathcal{L}(I, t, f_s, f_c) / \partial \theta_c.
\end{split}
\end{equation}

The pre-trained CLIP model is fixed during the update.

\subsection{One-shot learning mode}
\label{section:One-shot learning for each prompt}

Fig.~\ref{fig:method_oneshot} illustrates the proposed one-shot vector field generation method. Different from the iterative optimization mode, we train a UNet-based encoder-decoder network to directly predict the vector flow field $U_s$ and $U_c$.The process is defined as follows:
\begin{equation}
\begin{split}
    U_s = \mathcal{D}(\mathcal{E}(I,t)),\\
    \theta_c = \mathcal{H}(\mathcal{E}(I,t)),\\
    U_c = f_c(x,y|\theta_c),
\end{split}
\end{equation}
where $\mathcal{E}$ and the $\mathcal{D}$ denote the encoder and the decoder of the UNet, respectively. $\mathcal{H}$ is a hyper network combined by several MLP layers. $\mathcal{H}$ takes the encoding result $\mathcal{E}(I,t)$ as input and predict the parameters $\theta_c$ of our color flow field $f_c$.

Text prompt $t$ is mapped to the intermediate layers of the UNet via a cross-attention layer~\cite{vaswani2017attention} implementing $Attention(Q,K,V)=softmax(\frac{QK^T}{\sqrt{d}}) \cdot V$, with:
\begin{equation}
\begin{split}
    Q=W_Q^{(i)}\cdot \phi_i(z_i), K=W_K^{(i)} \cdot E(t), V=W_V^{(i)} \cdot E(t),
\end{split}
\end{equation}
where $E$ is a pre-trained CLIP text encoder, $\phi_i(z_i)$ denotes a flattened intermediate representation of the $i-th$ Unet feature map and $W_Q^{(i)},W_K^{(i)},W_V^{(i)}$ are learnable projection matrices\cite{jaegle2021perceiver, vaswani2017attention}.

We train the above network $\mathcal{D}$, $\mathcal{E}$, and $\mathcal{H}$ on the FFHQ~\cite{karras2019style} dataset. Without preparing any paired data (text, vector flow field), we use the pre-trained CLIP model to evaluate the semantic consistency between the input text prompts and the final output image $I^\prime$ ($I^\prime = T_{wp}(I, U_s, U_c)$). The whole loss term is equal to Eq.~(\ref{eq:opt}).

\begin{figure*}[t]
    \centering
    \includegraphics[width=\linewidth]{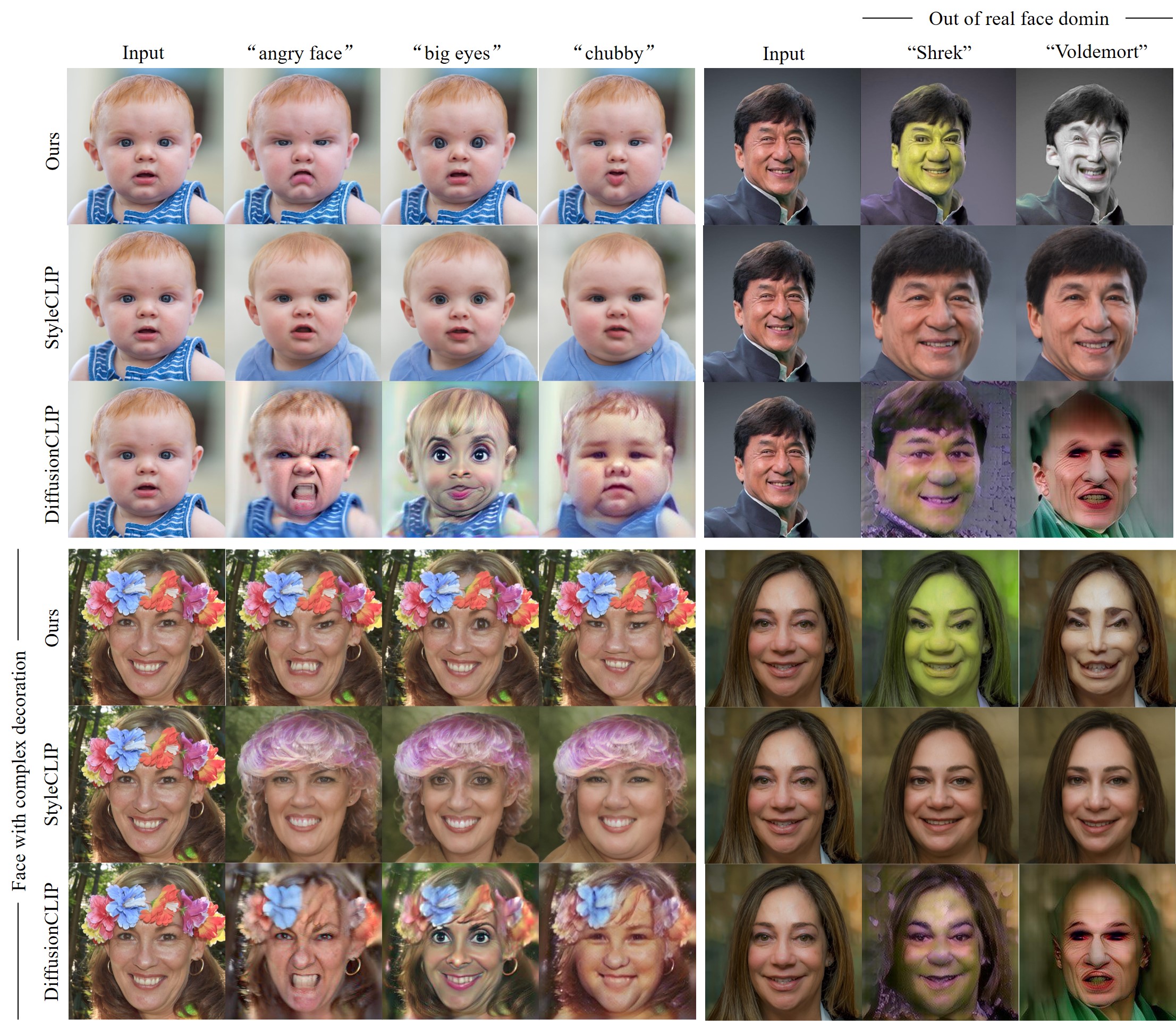}
    \caption{Visual comparison result with GAN-based SOTA method StyleCLIP~\cite{patashnik2021styleclip} and diffusion-based SOTA method DiffusionCLIP~\cite{Kim_2022_CVPR} using different text prompts. Our method can generate results out of the real human-face domain, e.g., if we use 'Shrek' as the text prompt, our method can not only modify the shape of the face but also change its skin to green. Our method also has high face identity consistency, which only edits properties that must be changed.}
    \label{fig:comparision}
\end{figure*}

\subsection{Real-time face video editing}

Our method can also be easily extended to the face editing task in videos. We only need to predict the vector field for the first video frame, then the vector flow fields for the subsequent frames can be generated by Homography transformation in real-time. The Homography matrix is calculated based on the facial landmarks tracked by the ERT method~\cite{kazemi2014one}. Since our method is based on successive explicit transformations, this enables our results to have good inter-frame consistency, while previous GAN-based methods are hard to achieve consistent results.

\subsection{Implementation Details}

{\bf INR network architecture.} Our INR network for the spatial flow field is configured as an MLP with 2 hidden layers. Each hidden layer has 32 neurons. The input layer has 18 neurons to take in the coordinates after position encoding, while its output layer has 2 neurons to predict the $\delta x$ and $\delta y$ to generate the warping vector flow field $U_s$.

{\bf Color editing.} Instead of computing the color changes directly in RGB space, we first transform the color to the HSV (Hue, Saturation, Value) space and the $U_c$ represents the value changes in the HSV space. 
% The color transform module consists of two single convolution layers with kernel size $1\times 1$. We use linear combinations of the original colors of the image to calculate the changes in the Hue and Saturation channels while keeping the Value channel unchanged.

{\bf Loss functions.} In addition to the CLIP similarity loss, we also use smoothness loss $\mathcal{L}_{sm}$ to ensure the continuity of the vector flow field and color loss $\mathcal{L}_{color}$ to control color changes. Regularization loss $\mathcal{L}_{reg}$ is necessary for the implicit representation approach to limit the complexity of the vector flow field and avoid over-distortion. The losses are defined as follows:
\begin{equation}
    \begin{split}
        \mathcal{L}_{sm}(U_s) &= \mathbb{E}_{x,y}\{ (\frac{\partial U_s(x,y)}{\partial x})^2 + (\frac{\partial U_s(x,y)}{\partial y})^2 \}, \\
    \mathcal{L}_{reg}(U_s) &= \mathbb{E}_{x,y}\{ U_s(x,y)^2 \}, \\
    \mathcal{L}_{color}(U_c) &= \mathbb{E}_{x,y}\{ (T_{wp}(I, U_c) - I)^2 \},\\
    \mathcal{L}_{id}(I, I^\prime) &= \mathbb{E}_{x,y}\{ 1-<\mathcal{F}(I),\mathcal{F}(I^{\prime})> \},
    \end{split}
\end{equation}
where $\mathcal{F}$ is a pre-trained face recognition model "Light CNN-29 v2"\cite{wu2018light}. We extract the 256-d facial embeddings of the input face and edited face and then compute the cosine distance between them as their similarity. We set $\lambda_{clip}$ in $[10,50]$, $\lambda_{sm} $in $[10,50]$, $\lambda_{reg}$ in $[0.1,1]$, $\lambda_{color}$ in $[10,100]$, and $\lambda_{id}$ in $[0,1]$.

{\bf Training details.} Our method is implemented using the Pytorch~\cite{pytorch} framework. In the iterative optimization mode, we use Adam optimizer~\cite{kingma2014adam} to update the parameters of the vector flow field. We set the learning rate to 1e-2, and the maximum number of iterations to 3000. We reduce the learning rate to 1/2 every 1000 epochs. In the one-shot learning mode, we fix the CLIP encoders and train the network $\mathcal{D}$, $\mathcal{E}$, and $\mathcal{H}$. The learning rate is set to 1e-4. The learning rate decays to 1/2 every 10 epochs and the training stops after 100 epochs. In the two modes above, we use RandAugment~\cite{NEURIPS2020_d85b63ef} to enhance the training robustness. The RandAugment randomly chooses augmentation transformations with random magnitude to augment the input images of the CLIP encoder. We find this method can help improve the stability of the CLIP model more efficiently. 

% or This method has been proven by [xxx] to help the CLIP model identify input images more efficiently.

\section{Experimental Analysis}

\begin{table*}
\centering
\caption{Quantitative comparison of different methods}\label{tab:Quantitative}

\begin{threeparttable}
\begin{tabular}{l|l|cccc|cc}
\toprule
    & & \multicolumn{4}{c}{common human expressions (In domain)} & \multicolumn{2}{c}{movie characters (Out of domain)} \\
\midrule
\textmd{metric} &\textmd{method} & \textmd{"angry face"}  & \textmd{"smiling face"}  &  \textmd{"big eyes"} &\textmd{"chubby"} &\textmd{"Shrek"} &\textmd{"Voldemort"}\\
\midrule
    \multirow{5}{*}{\textmd{$S_{CLIP}$~\cite{radford2021learning}$\uparrow$}}
    & \textmd{StyleCLIP~\cite{patashnik2021styleclip}}& $0.2281$ & $0.2337$ & $0.2194$ & $0.2357$ & $0.2060$ & $0.2099$\\
    & \textmd{DiffusionCLIP~\cite{Kim_2022_CVPR}}& $0.2589$ & $0.2504$ & $0.2548$ & $\textbf{0.2575}$ & $0.2695$ & $\textbf{0.3122}$\\
    & \textmd{Ours-T\tnote{1}}& $\textbf{0.2747}$ & $\textbf{0.2530}$ & $\textbf{0.2678}$ & $0.2451$ & $\textbf{0.2920}$ & $0.2876$\\
    & \textmd{Ours-I\tnote{2}}& $0.2531$ & $0.2421$ & $0.2440$ & $0.2340$ & $0.2882$ & $0.2683$\\
    & \textmd{Ours-O\tnote{3}}& $0.2669$ & $0.2379$ & $0.2610$ & $0.2383$ & $0.2837$ & $0.2622$\\
    % & \textmd{Ours-Oneshot}& $1$ & $1$ & $1$ & $1$ & $1$ & $1$\\
\midrule
    \multirow{5}{*}{\textmd{FID~\cite{Seitzer2020FID}$\downarrow$}}
    & \textmd{StyleCLIP~\cite{patashnik2021styleclip}}& $55.47$ & $59.18$ & $60.87$ & $65.32$ & $55.61$ & $56.93$\\
    & \textmd{DiffusionCLIP~\cite{Kim_2022_CVPR}}& $88.58$ & $97.70$ & $95.04$ & $97.43$ & $103.46$ & $116.77$\\
    & \textmd{Ours-T}& $10.94$ & $9.73$ & $13.59$ & $8.85$ & $\textbf{43.41}$ & $34.69$\\
    & \textmd{Ours-I}& $\textbf{6.33}$ & $\textbf{6.42}$ & $\textbf{7.04}$ & $\textbf{6.82}$ & $47.22$ & $\textbf{25.65}$\\
    & \textmd{Ours-O}& $32.75$ & $37.33$ & $42.19$ & $35.70$ & $69.32$ & $46.14$\\
     % & \textmd{Ours-Oneshot}& $1$ & $1$ & $1$ & $1$ & $1$ & $1$\\
\midrule
    \multirow{5}{*}{\textmd{ID~\cite{deng2019arcface}$\uparrow$}}
    & \textmd{StyleCLIP~\cite{patashnik2021styleclip}}& $0.7171$ & $0.7586$ & $0.6906$ & $0.6727$ & $\textbf{0.7803}$ & $0.7476$\\
    & \textmd{DiffusionCLIP~\cite{Kim_2022_CVPR}}& $0.5790$ & $0.5665$ & $0.4661$ & $0.5936$ & $0.5876$ & $0.2914$\\
    & \textmd{Ours-T}& $0.8010$ & $0.9133$ & $0.8824$ & $0.8509$ & $0.6825$ & $0.6778$\\
    & \textmd{Ours-I}& $\textbf{0.9100}$ & $\textbf{0.9312}$ & $\textbf{0.9535}$ & $\textbf{0.9065}$ & $0.6732$ & $\textbf{0.7638}$\\
    & \textmd{Ours-O}& $0.6560$ & $0.7365$ & $0.7589$ & $0.6822$ & $0.6595$ & $0.6051$\\
     % & \textmd{Ours-Oneshot}& $1$ & $1$ & $1$ & $1$ & $1$ & $1$\\
\bottomrule
\end{tabular}

 \begin{tablenotes}
        \footnotesize
        \item[1] Ours-T means our iterative optimization mode that explicitly represents the flow vectors with rasterized tensor
        \item[2] Ours-I means our iterative optimization mode that implicitly parameterizes the flow vectors by the INR network
        \item[3] Ours-O means our one-shot learning mode
      \end{tablenotes}
\end{threeparttable}

\end{table*}

\subsection{Comparison with other methods}

We compare our model with other classic text-driven face editing methods, including the GAN-based method StyleCLIP~\cite{patashnik2021styleclip} (Official implementation, global direction mode) and diffusion-based method DiffusionCLIP~\cite{Kim_2022_CVPR} (Official implementation, official fine-tune method).

We analyze the model performances under different types of text prompts. Our text prompts include common human expressions (such as ``angry face''), whose desired outputs are in the human face domain. We also use some complex text descriptions from well-known film characters (such as ``Shrek''), whose desired outputs are out of the common face domain.

\textbf{Visual comparison.} Fig.~\ref{fig:comparision} shows the visual comparison results. Our results are generated by the iterative optimization mode and directly optimize the rasterized tensors. For common human expressions, all three methods are able to generate semantic consistency results. However, when editing faces with complex decoration details, StyleCLIP and DiffusionCLIP both failed to reconstruct these details in the output image. StyleCLIP even causes identity error results, which reconstructs the lady's headwear flowers into colored curled hair. Part of the reason is that inversing the face into the latent space is hard to be completely accurate. Our method only changes the face properties that need to be changed. For text prompts from well-known film characters, StyleCLIP fails and makes very little change to the face. This is because the desired outputs are out of the pre-trained GAN's domain, which is the main limitation of GAN-inversion-based methods. DiffusionCLIP edits the face to look like the movie character according to the text prompt, but its performance of maintaining the original facial identity is low. While maintaining the consistency of the face, our method edits the input face to have the main characteristics of the movie character ——Shrek's big nose, big round mouth, and green face, and Voldemort's flat nose, pale and ferocious face, etc.

\textbf{Quantitative comparison.} We make the quantitative comparison on the FFHQ~\cite{karras2019style} test dataset, detailed parameters settings are in our appendix. For quantitative comparison metrics, We evaluate the semantic consistency $S_{CLIP}$ by the CLIP~\cite{radford2021learning} similarity, the visual quality by Fr\'{e}chet Inception Distance (FID) ~\cite{Seitzer2020FID}, and face identity by pre-trained face recognition model Arcface~\cite{deng2019arcface}. Tab.~\ref{tab:Quantitative} shows the quantitative results. For the FID metric and ID metric, our method has always outperformed the comparative methods by a wide margin, which shows our method has better image quality and face identity consistency. (In addition to StyleCLIP's ID metric when inputting movie character names, just because it even doesn't change the faces.) This is because our face edit results are based on the original image information, the encoding-decoding process is not required. For the semantic consistency $S_{CLIP}$, our method is always better than StyleCLIP and also outperforms DiffusionCLIP in most text. 

% From the results, we can see that the proposed method can understand the text prompts and generate face-editing results with consistent semantics between the result and the text. 
% Our proposed method can understand the names of celebrities or characters, i.e. Red Skull, and also can distinguish emotions. In addition, the color can be also properly edited under prompting. For example, Shrek is a famous animated character with green skin. If we use 'Shrek' as the text prompt, the proposed method can not only modify the shape of the face but also change its color to green.

\begin{figure}
    \centering{\includegraphics[width=1.0\linewidth]{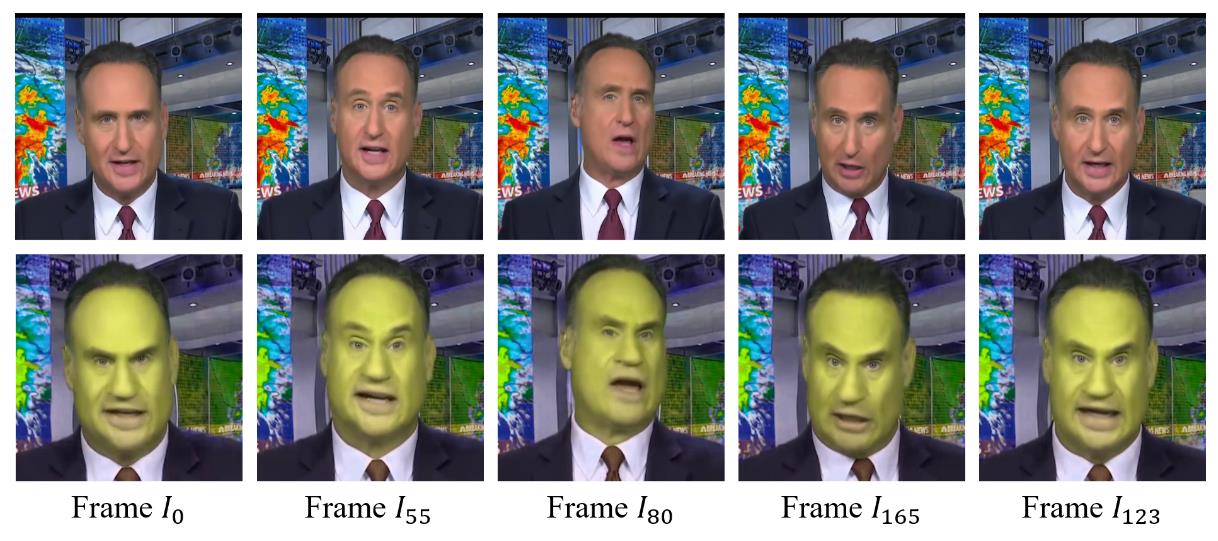}}\\
    \caption{Real-time face video editing results. The 1st - 2nd rows show the original images and face editing results on different frames, respectively.}    
    \label{fig:video_warping_results}
\end{figure}

\subsection{Real-time face video warping}

In Fig.~\ref{fig:video_warping_results}, we show our real-time face editing result on a video from a weather forecast TV program with the text prompt "Green Hulk". The results show that our method has good interframe consistency.

\begin{figure}
    \centering{\includegraphics[width=1.0\linewidth]{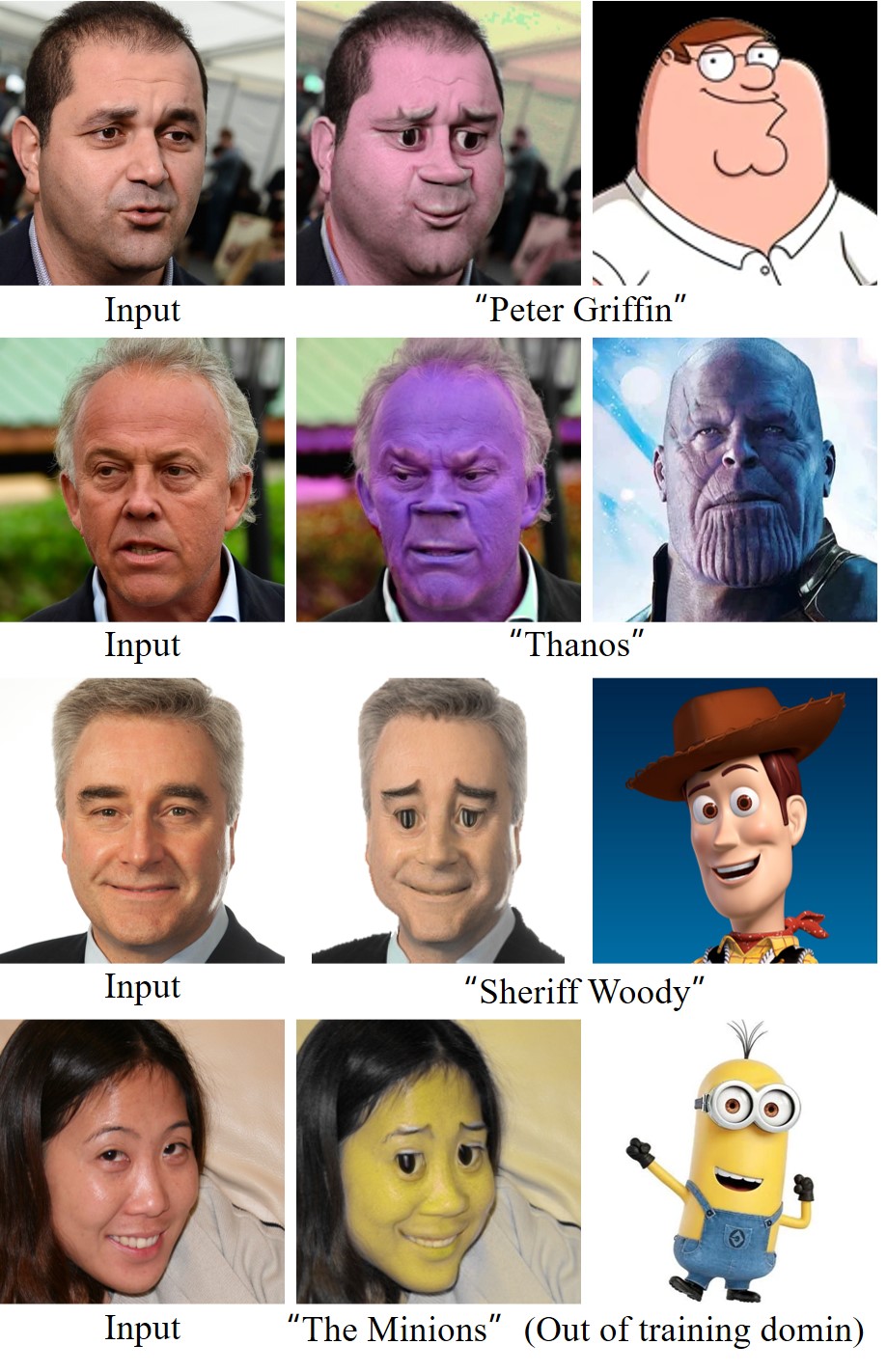}}\\
    \caption{Visualization results of our one-shot learning mode. Note that the text prompt "The Minions" is not included in our training set, but our network has the ability to generalize, generating minion-style images with big round eyes and yellow faces. }
    \label{fig: oneshot-visual}
\end{figure}
\subsection{Visualization results of our one-shot learning mode}
Fig.~\ref{fig: oneshot-visual} is the visualization results of our one-shot learning mode. We show some text prompts from well-known anime characters to show that our approach is capable of various forms of facial deformation. The network can be generalized to unseen text prompts because it can analyze the character's appearance from the clip embedding space. Our deformation-based method makes the input image have the primary features of cartoon characters, such as skin color, facial shape, and line features while maintaining identity consistency.

\begin{figure}
    \centering{\includegraphics[width=1.0\linewidth]{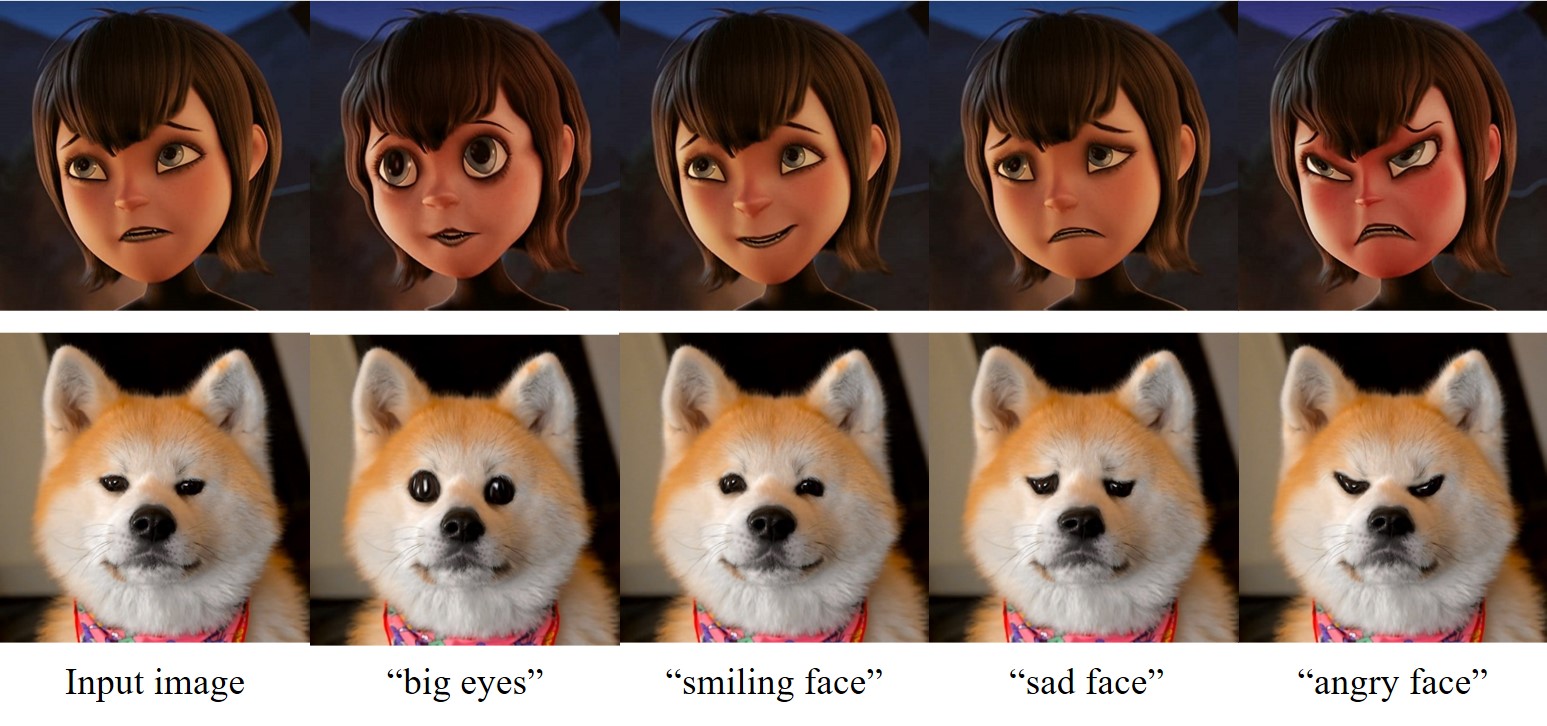}}\\
    \caption{Face editing results of cartoon faces and animal faces. Generated by our iterative optimization mode.}
    \label{fig: cartoon}
\end{figure}

\subsection{Editing cartoon faces and animal faces}
Because our approach does not rely on any per-category pre-training model, our method can be directly applied to cartoon faces or animal faces, as shown in Fig.~\ref{fig: cartoon}.
 
% The results show that our method can stably and smoothly generate faces corresponding to a given text prompt. 
% We recommend reviewers check out our supplementary material for animated demonstrations.

\begin{figure}
    \centering{\includegraphics[width=0.8\linewidth]{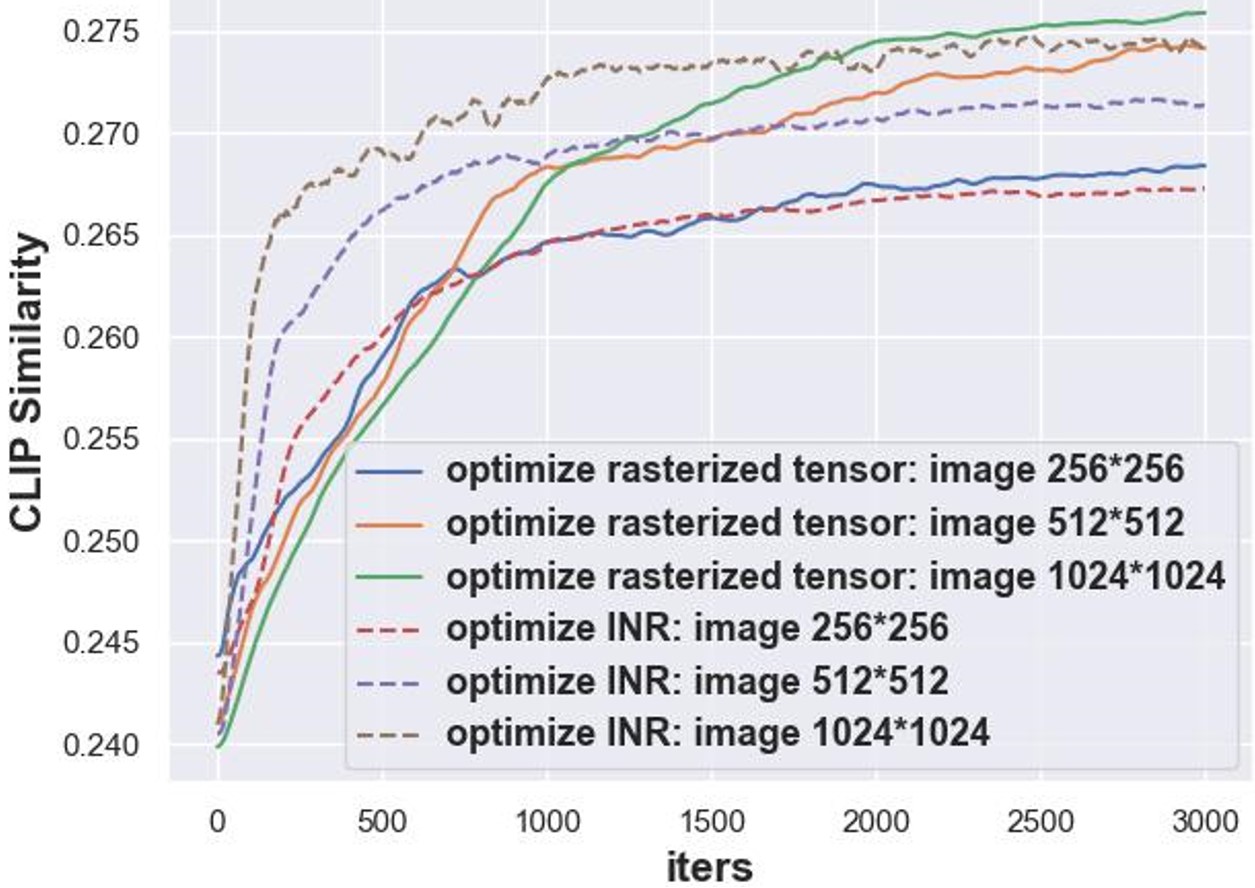}}\\
    \caption{Semantic accuracy of different vector flow field generating methods in different image resolutions. }
    \label{fig: flow_int_compare_loss}
\end{figure}

\begin{figure}
    \centering{\includegraphics[width=1.0\linewidth]{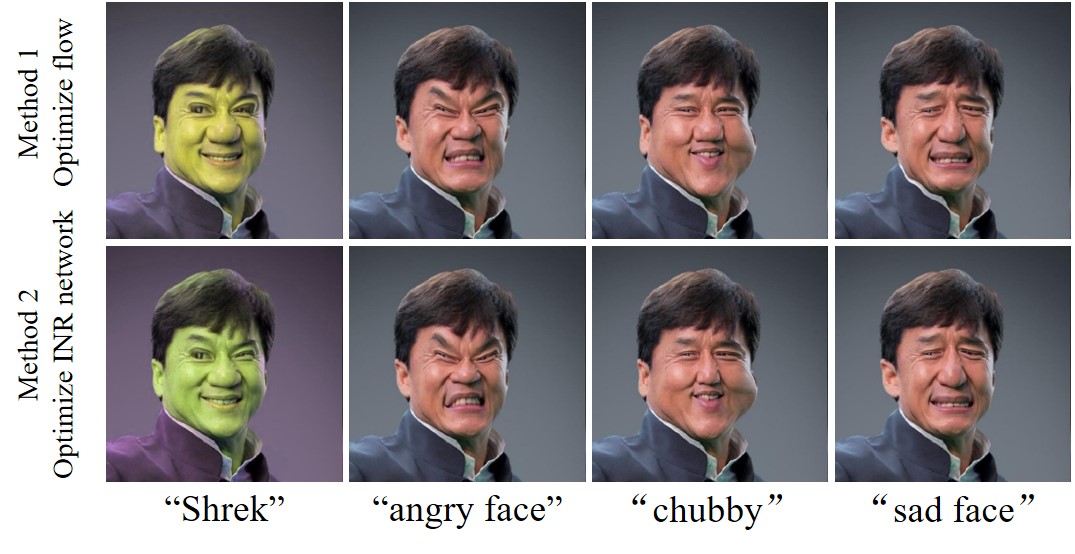}}\\
    \caption{Face editing visualization results of different vector flow field generating methods. 
    % The first row is generated by explicitly optimizing the rasterized flow. The second row is generated by optimizing the INR network to implicitly parameterize the flow vectors.
    }
    \label{fig: flow_int_compare_vis}
\end{figure}

\subsection{Comparison of different vector flow fields generating methods}

We represent the flow vectors in two approaches: 1) explicitly represent the flow vectors with rasterized tensor, and 2) implicitly parameterize the flow vectors as continuous, smooth, and resolution-agnostic neural fields. Here we compare the parameter numbers, accuracy, and speed of these two approaches. 

The parameter number in the second method~(1.7k, in our designed architecture) is far less than the number of the parameters in the first method~(524k for a 512$\times$512 pixel image). We use various different-resolution images and record the CLIP Similarity in each optimization step of these two approaches, representing semantic accuracy. As shown in Fig.~\ref{fig: flow_int_compare_loss}, high-resolution images tend to achieve higher semantic accuracy after editing, but cost more iteration steps. The second method has a faster convergence speed, which can reduce iteration steps significantly. Since the second method is to update the flow field in the parameter weight space of the INR, the flow field can be globally optimized and has a faster convergence speed. However, as the number of iterations increases, the semantic accuracy of the first method continues to increase and exceeds that of the second method, which is also illustrated by the visualization results in Fig.~\ref{fig: flow_int_compare_vis}. The facial warping result of the first method is more refined.

\subsection{Detailed structure of our one-shot network}
The encoder $\mathcal{E}$ and the decoder $\mathcal{D}$ in our one-shot network are similar to the Unet model used in~\cite{rombach2021highresolution}. In the encoder $\mathcal{E}$, we down-sample the input image 3 times. In each down-sample level, we set 2 residual blocks. The number of channels in each down-sample level is (64, 128, 128, 256), respectively. We use a stride-2 convolution layer to achieve the down-sample. In the last down-sample level (where the feature map has the minimum size), we map the text encoding result to the UNet layer by the muti-head cross-attention mechanism. The number of attention heads is set to 8 and the channel width per attention head is set to 32. The decoder $\mathcal{D}$ has a symmetric structure with the encoder. It uses the nearest sampling method to make upsampling.
We use the encoding result $\mathcal{D}(I, t)$to feed forward into the hyper-network $\mathcal{H}$ and predict the weight and parameters of the color flow field $f_c$. The hyper-network $\mathcal{H}$ is an MLP with two hidden layers. Each hidden layer has 256 neurons and uses the ReLU activate functions.

\subsection{Evaluations of our INR network in the iterative optimization mode}
\begin{figure}
    \centering{\includegraphics[width=1.0\linewidth]{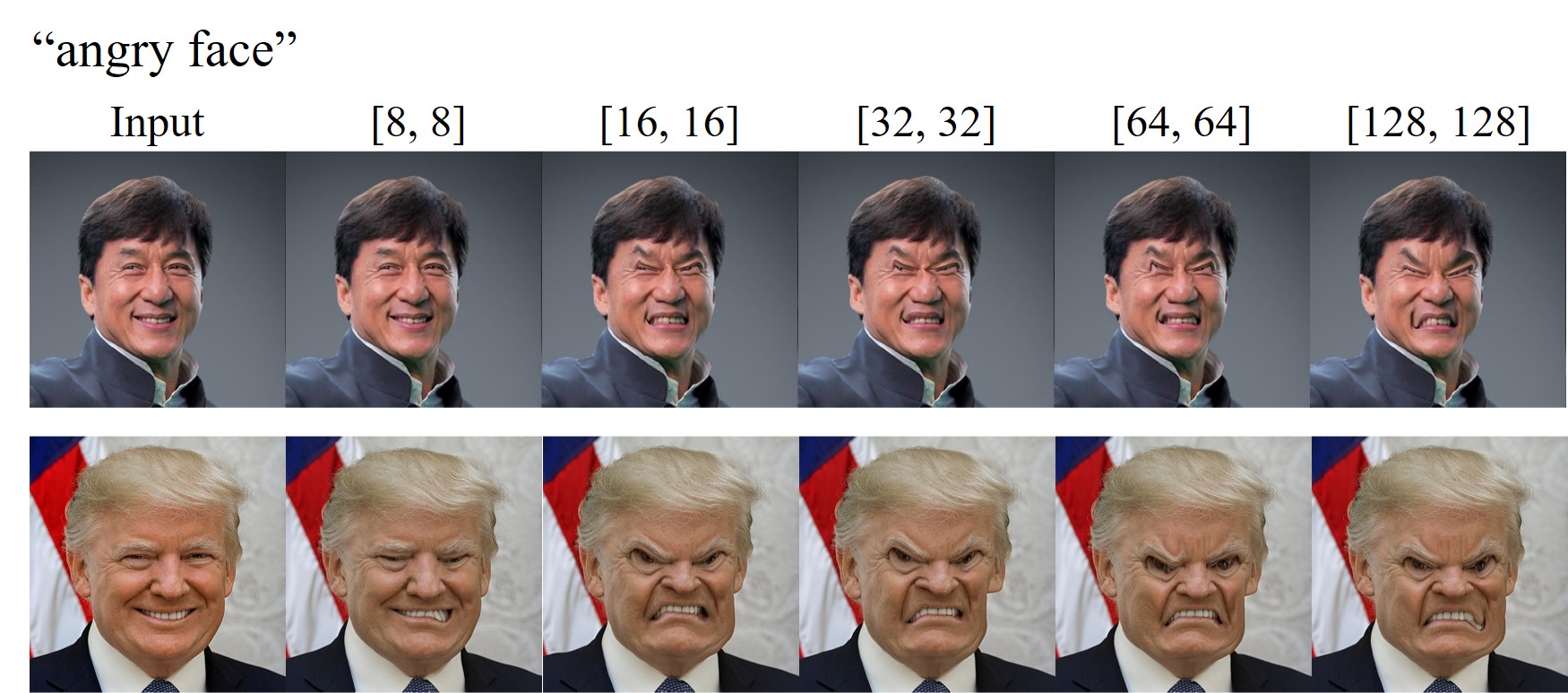}}\\
    \caption{The number of neurons per hidden layer in the network could influence the amplitude of warping under the same optimization iterations.}
    \label{fig:number_of_neurons}
\end{figure}

\begin{figure}
    \centering{\includegraphics[width=1.0\linewidth]{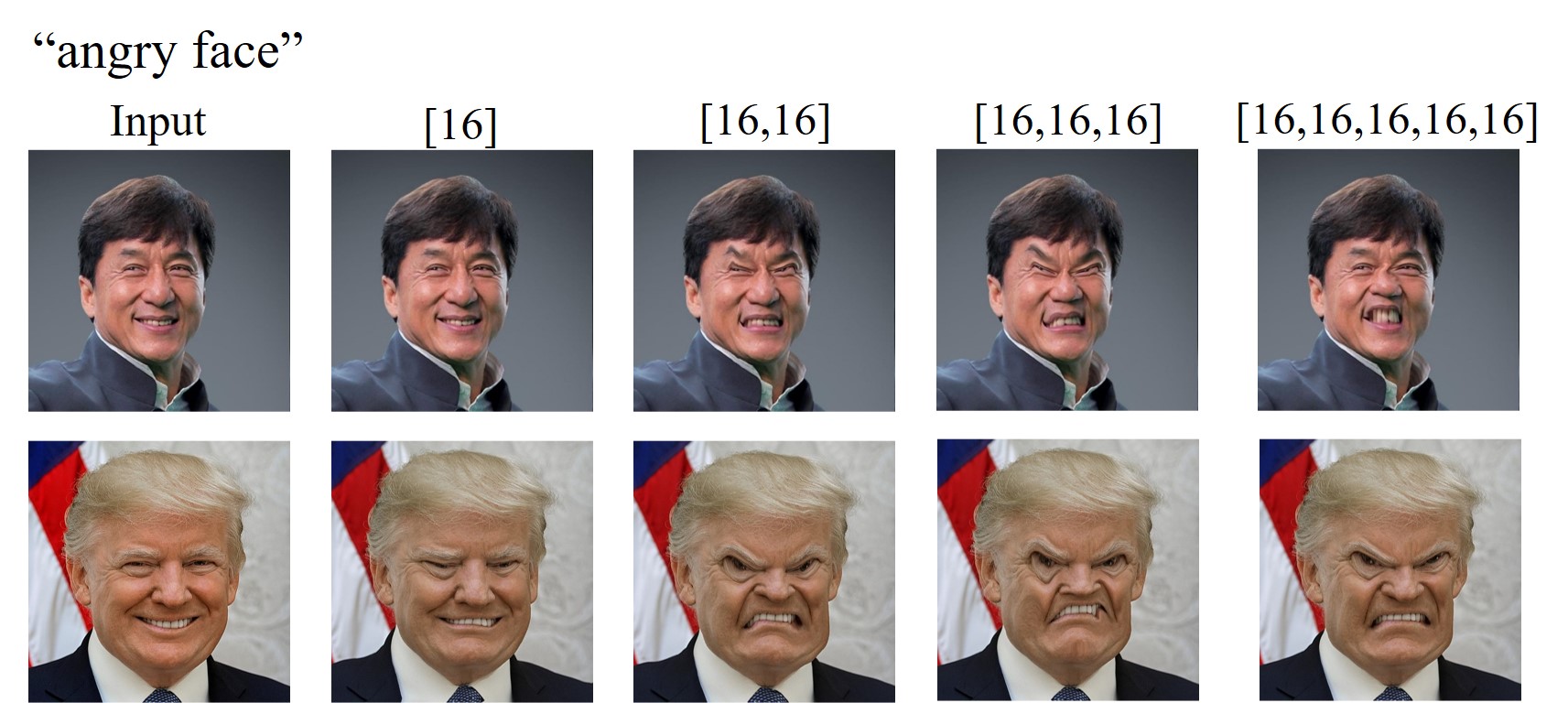}}\\
    \caption{The number of hidden layers in the network could influence the amplitude of warping under the same optimization iterations.}
    \label{fig:number_of_layers}
\end{figure}

\textbf{Number of neurons in INR network} The ability to implicitly represent the vector flow field mainly depends on the structure of our INR network. We adjust the number of neurons and hidden layers in the INR network to obtain a network that can represent complex warping and be optimized faster. As shown in Fig.~\ref{fig:number_of_neurons} and Fig.~\ref{fig:number_of_layers} respectively. In our experiments, we use "angry face" as the text prompt.

First, we investigate the effect of the number of neurons on the performance of the optimization. On the basis of two hidden layers, we set a different number of neurons in each hidden layer. Fig.~\ref{fig:number_of_neurons} shows the experiment results. It can be observed that as the number of neurons increases, the amplitude of warping becomes larger. When each hidden layer has 8 neurons, the image undergoes only a small deformation, especially at the mouth and eyes, which is far from the "angry face". However, when the number of neurons per layer is larger than 64 or even 128, the image will be overshaped, especially overfitted in the eyes and mouth areas. Therefore, the network has the best performance when each hidden layer has 32 neurons, which also reduces the number of parameters and ensures a faster optimization speed.

Next, we investigate the effect of the number of hidden layers on the performance of the optimization. Fig.~\ref{fig:number_of_layers} shows the experiment results. It can be observed that using more hidden layers can significantly fit more complex deformations and increase the amplitude of warping. However, if the setting is not reasonable, the network will also cause overfitting.

\begin{figure}
    \centering{\includegraphics[width=1.0\linewidth]{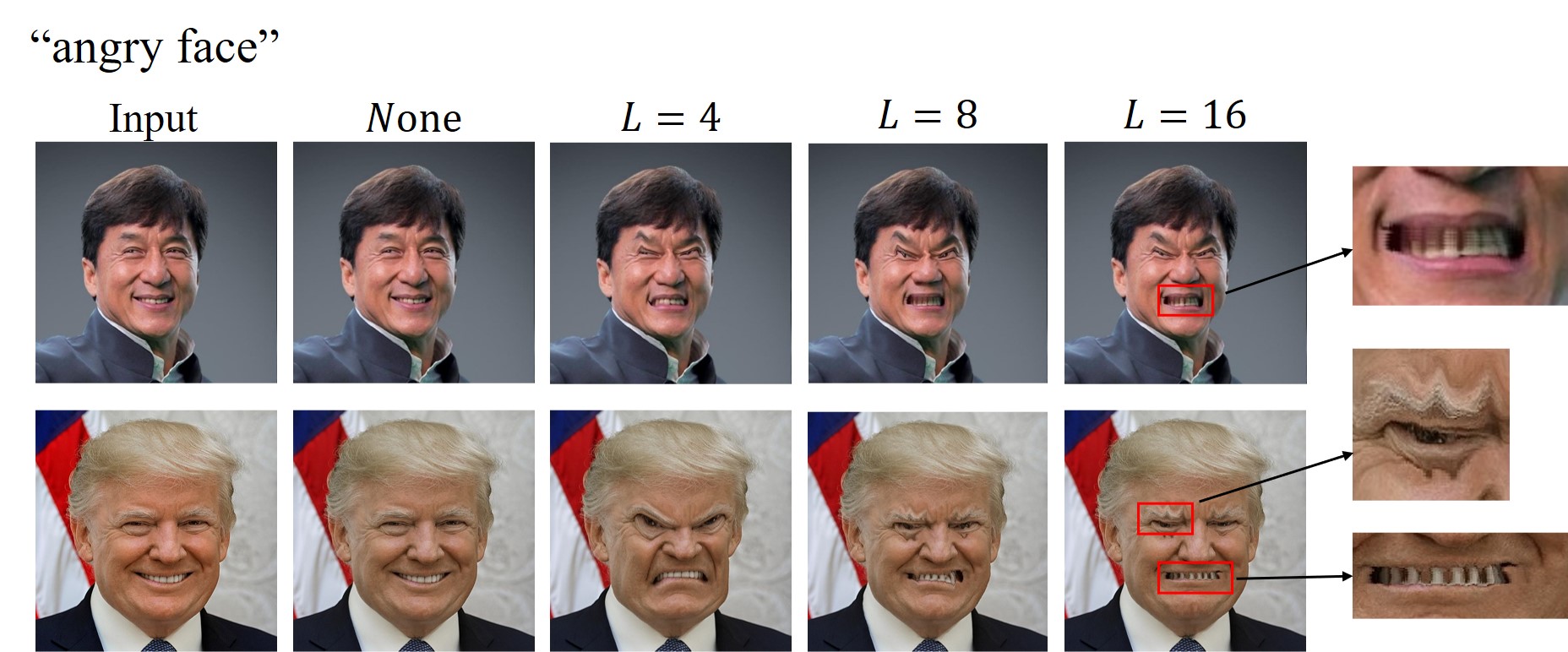}}\\
    \caption{The number of position encoding layers influences the final performance of our model. The larger $L$ may result in local wrinkle and blur as indicated by the red box.}
    \label{fig:position_encoding}
\end{figure}

\begin{table*} %[h]\small
\centering
\caption{Details quantitative experiment settings.}
\begin{tabular}{l|ccccccc}
    \toprule    
      & change color & $\lambda_{clip}$ & $\lambda_{sm}$ & $\lambda_{color}$ & $\lambda_{id}$ & $\lambda_{reg}$ & blur kernel size\\
    \midrule
    & \multicolumn{7}{c}{Optimize rasterized tensor}\\
    \midrule
     "angry face"   & false & $10.0$ & $10.0$  & - & $0.1$ & $0.0$  & $71$\\
     "smiling face" & false & $10.0$ & $10.0$  & - & $0.1$ & $0.0$  & $71$\\
     "big eyes"     & false & $10.0$ & $10.0$  & - & $0.5$ & $0.0$  & $51$\\
     "chubby"       & false & $30.0$ & $10.0$  & - & $0.2$ & $0.0$  & $91$\\
     "Shrek"        & true & $10.0$ & $10.0$  & $20.0$ & $0.1$ & $0.0$  & $51$\\
     "Voldemort"    & true & $30.0$ & $10.0$  & $50.0$ & $0.1$ & $0.0$  & $71$\\
    \midrule
    & \multicolumn{7}{c}{Optimize the INR network} \\
    \midrule
     "angry face"   & false & $10.0$ & $10.0$  & - & $0.1$ & $0.1$  & -\\
     "smiling face" & false & $10.0$ & $10.0$  & - & $0.1$ & $0.1$  & -\\
     "big eyes"     & false & $10.0$ & $10.0$  & - & $0.5$ & $0.5$  & -\\
     "chubby"       & false & $30.0$ & $10.0$  & - & $0.1$ & $0.1$  & -\\
     "Shrek"        & true & $10.0$ & $10.0$  & $20.0$ & $0.1$ & $0.1$  & -\\
     "Voldemort"    & true & $30.0$ & $10.0$  & $50.0$ & $0.1$ & $0.1$  & -\\
     \midrule
    & \multicolumn{7}{c}{Oneshot learning mode} \\
    \midrule
     -   & true & $10.0$ & $10.0$  & $10.0$ & $0.1$ & $0.0$  & $51$\\
    \bottomrule
\end{tabular}\label{tab:exp_config}
\end{table*}

\textbf{Position encoding}
In order to better represent high-frequency variation in the vector flow field, we add high dimensional position encoding to enable the INR network to learn higher frequency details. We do the position encoding operation before the vector flow field coordinates are passed to the INR network. Formally, the encoding function can be defined as follows:

\begin{equation}\label{eq:position_encoding}
\begin{split} 
    \gamma (p) = (p, & sin(2^0 \pi p),  cos(2^0 \pi p), ... ,\\
                     & sin(2^{L-1}\pi p),  cos(2^{L-1} \pi p)),
\end{split} 
\end{equation}
where $p$ denotes the coordinates $(x,y)$. In our experiments, we set different $L$ to see the different fitting abilities of the INR network. We also use “angry face” as the text prompt. The results can be seen in Fig.~\ref{fig:position_encoding}. The results show that this encoding function can significantly improve performance. However, a larger $L$ may cause overfitting in some areas, like the mouth and eyes. So, in our future experiments, we set $L=4$ to enable stable and excellent performance.

\subsection{Detailed configurations of quantitative experiment settings}
In quantitative experiments, we evaluate our method using different hyperparameters for different text prompts. Our detailed settings are in Tab.~\ref{tab:exp_config}. Note that when we implicitly parameterize the flow vectors, it's unnecessary to use the Gaussian blur kernel to smooth the flow field. However, we should add a regular loss term to restrict the complexity of the vector flow field and avoid over-distortion. Detailed verification of these hyperparameters is in Sec.~\ref{sec:loss_weight}.

\begin{table*} %[h]\small
\centering
\caption{Comparison of our iterative optimization and one-shot learning.}\label{tab:compare_oneshot}

\begin{threeparttable}
\begin{tabular}{l|ccccc}
    \toprule    
      & training time & total inference time & inference time per iter & input image dependent \\
      \midrule
optimize-T\tnote{1} & - & 8.6 minutes & 0.17 seconds & yes \\
optimize-I\tnote{2} & - & 8.6 minutes & 0.17 seconds & yes \\
Oneshot & 86.7 hours & 0.96 seconds & - & no \\
    \bottomrule
\end{tabular}

\begin{tablenotes}
        \footnotesize
        \item[1] optimize-T means our method that explicitly represents the flow vectors with rasterized tensor
        \item[2] optimize-I means our method that implicitly parameterizes the flow vectors by the INR network
      \end{tablenotes}
\end{threeparttable}
\end{table*}

\begin{figure}
    \centering{\includegraphics[width=1.0\linewidth]{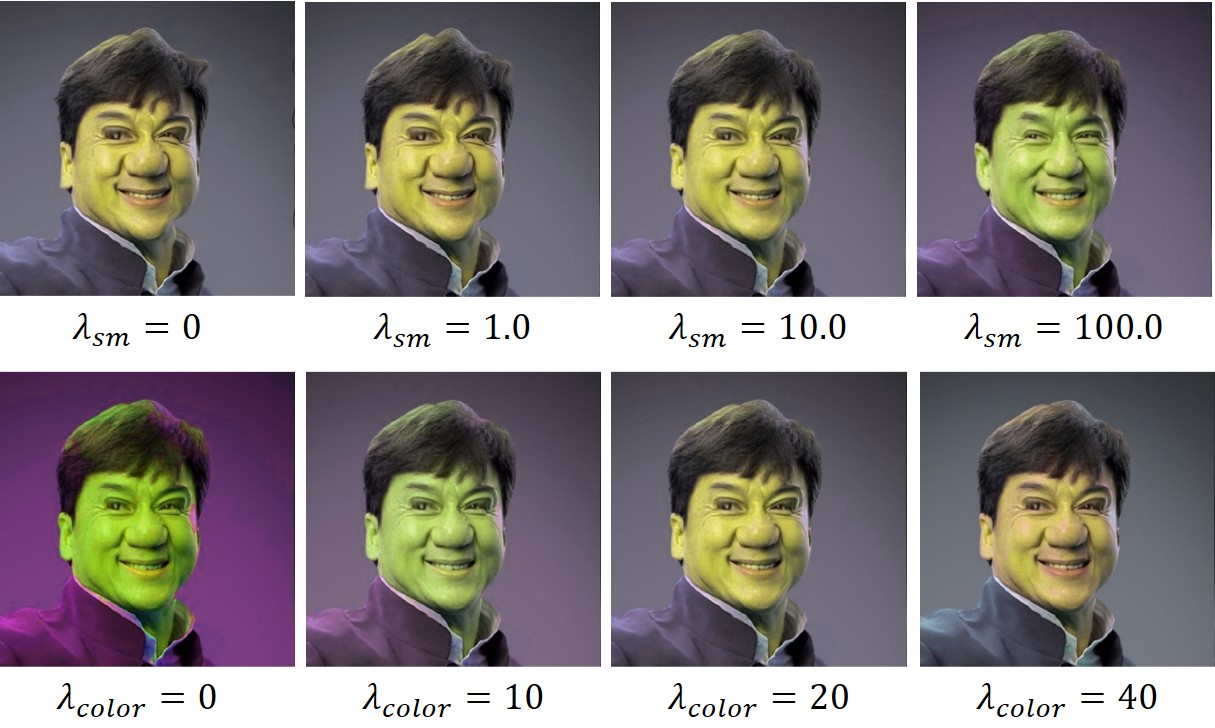}}\\
    \caption{The effectiveness of our smoothness loss and color loss.}
    \label{fig:smcolorloss}
\end{figure}

\begin{figure}
    \centering{\includegraphics[width=1.0\linewidth]{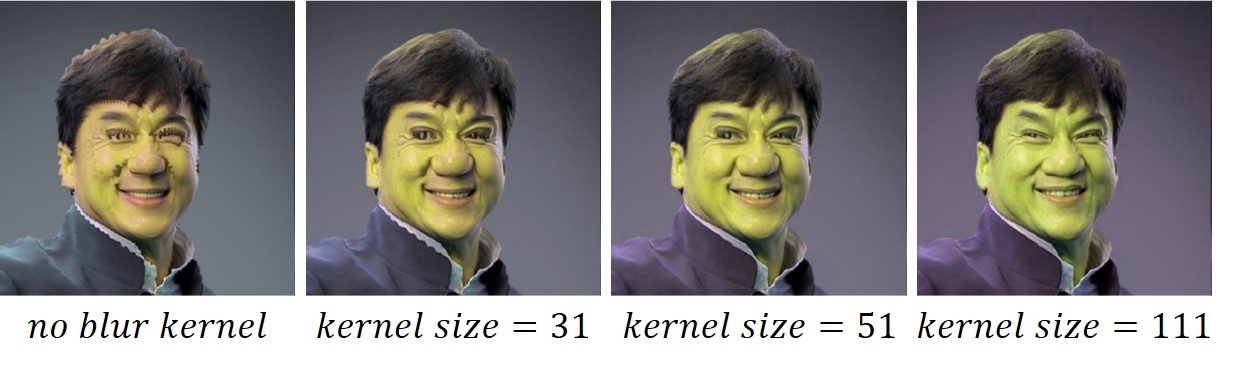}}\\
    \caption{The effectiveness of the Gaussian blur kernel when explicitly representing the vector flow field.}
    \label{fig:blur}
\end{figure}

\begin{figure}
    \centering{\includegraphics[width=1.0\linewidth]{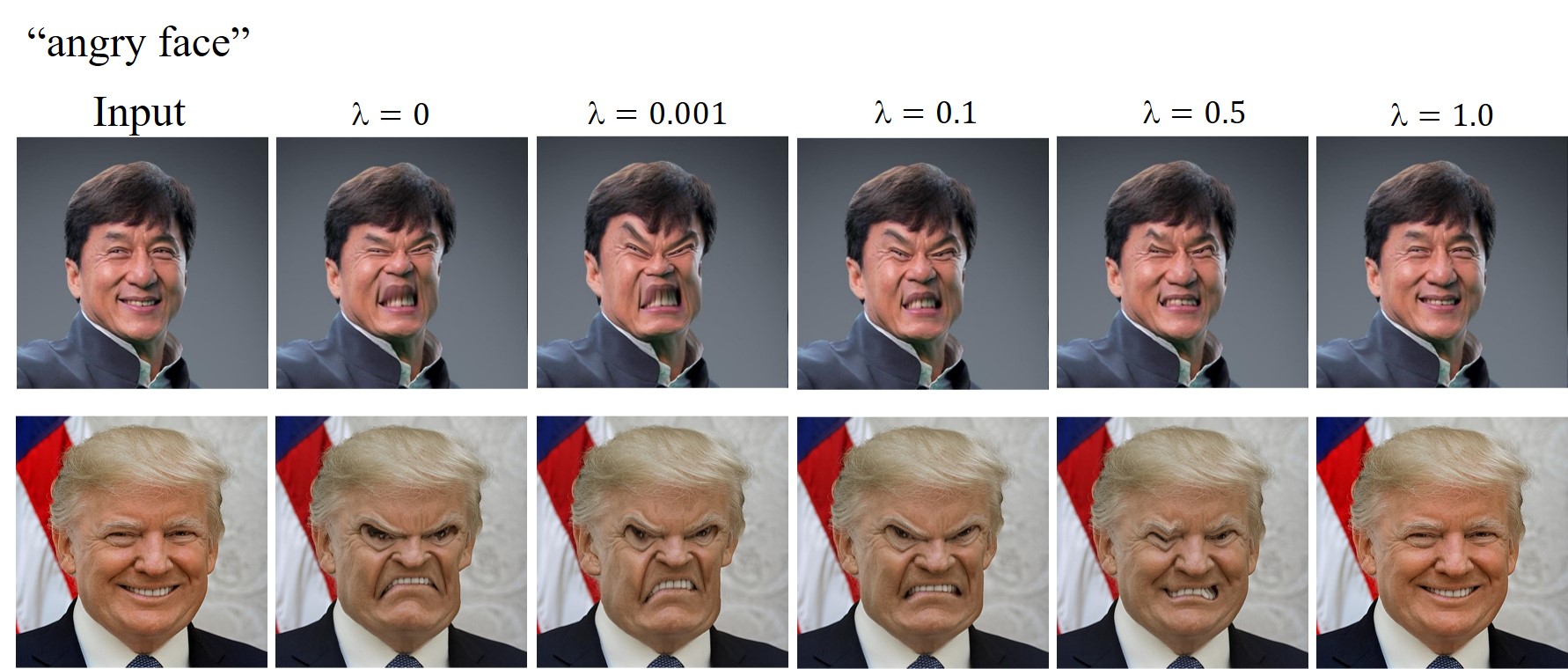}}\\
    \caption{The weight of the L2 norm regularization term influences the amplitude of warping.}
    \label{fig:L2_regularization}
\end{figure}

\subsection{Discusstions of our hyper-parameter settings}\label{sec:loss_weight}
\textbf{Smoothness loss and color loss} Smoothness loss $\mathcal{L}_{sm}$ and color loss $\mathcal{L}_{color}$ are necessary for our iterative optimization mode (both explicit and implicit flow vector representation) and our one-shot learning mode. Here we make an intuitive qualitative comparison of the effects of these loss terms and their loss weights. We use "Shrek" as the text prompt, set $\mathcal{L}_{clip}=10.0, \mathcal{L}_{sm}=10.0, \mathcal{L}_{color}=20.0, \mathcal{L}_{id}=0.1, \mathcal{L}_{reg}=0.0$ and set blur kernel size as 51. We make experiments in the iterative optimization mode with the explicit flow vector representation. We change the weight of one loss term in each experiment to evaluate its effect. The results are shown in Fig.~\ref{fig:smcolorloss}.

Smoothness loss $\mathcal{L}_{sm}$ is used to control the continuity of the vector flow field. Without smoothness loss, the face will produce an unsmooth contortion (Like the sharp corners of your hair). However, too large a smoothness loss weight will limit the necessary deformation of the face. Color loss $\mathcal{L}_{color}$ is used to limit colors to changing not too much. $\lambda_{color}$ determines if the color change is too dark or too light.

\textbf{Gaussian blur kernel size} Gaussian blur kernel is necessary for the explicit representation approach of the flow vectors. The qualitative experiment results are shown in Fig.~\ref{fig:blur}. It can be intuitively seen that after the removal of the Gaussian fuzzy kernel, the generated image will appear with many unsmooth high-frequency distortions. However, too large the large Gaussian blur kernel limits the necessary distortion (like Jackie Chan's nose), resulting in less vivid face editing.

\textbf{L2 norm regularization} Regularization loss term $\mathcal{L}_{reg}$ is necessary for the implicit representation approach of the flow vectors. As shown in Eq. (5) in the main text, hyper-parameter $\lambda_{reg}$ restricts the complexity of the vector flow field. We explore the role of regularization by adjusting the value of $\lambda_{reg}$. As shown in Fig.~\ref{fig:L2_regularization}.

In this experiment, we use "angry face" as the text prompt, and adjust the coefficients of the L2 regularization term. Increasing the weight of the L2 regularization term can effectively limit the magnitude of the image deformation: if $\lambda_{reg}=0$, i.e., the term is not added to the loss function, the nose and eyes will be over-deformed; if the setting is too large, it will increase the constraint on the image deformation: as shown in the last column, the results do not yet have such characteristics of "angry face". Therefore, a reasonable choice of $\lambda_{reg}$, approximately $[0.1, 0.5]$, can ensure both the appropriate amplitude of warping and optimization speed simultaneously.

\begin{figure}
    \centering{\includegraphics[width=1.0\linewidth]{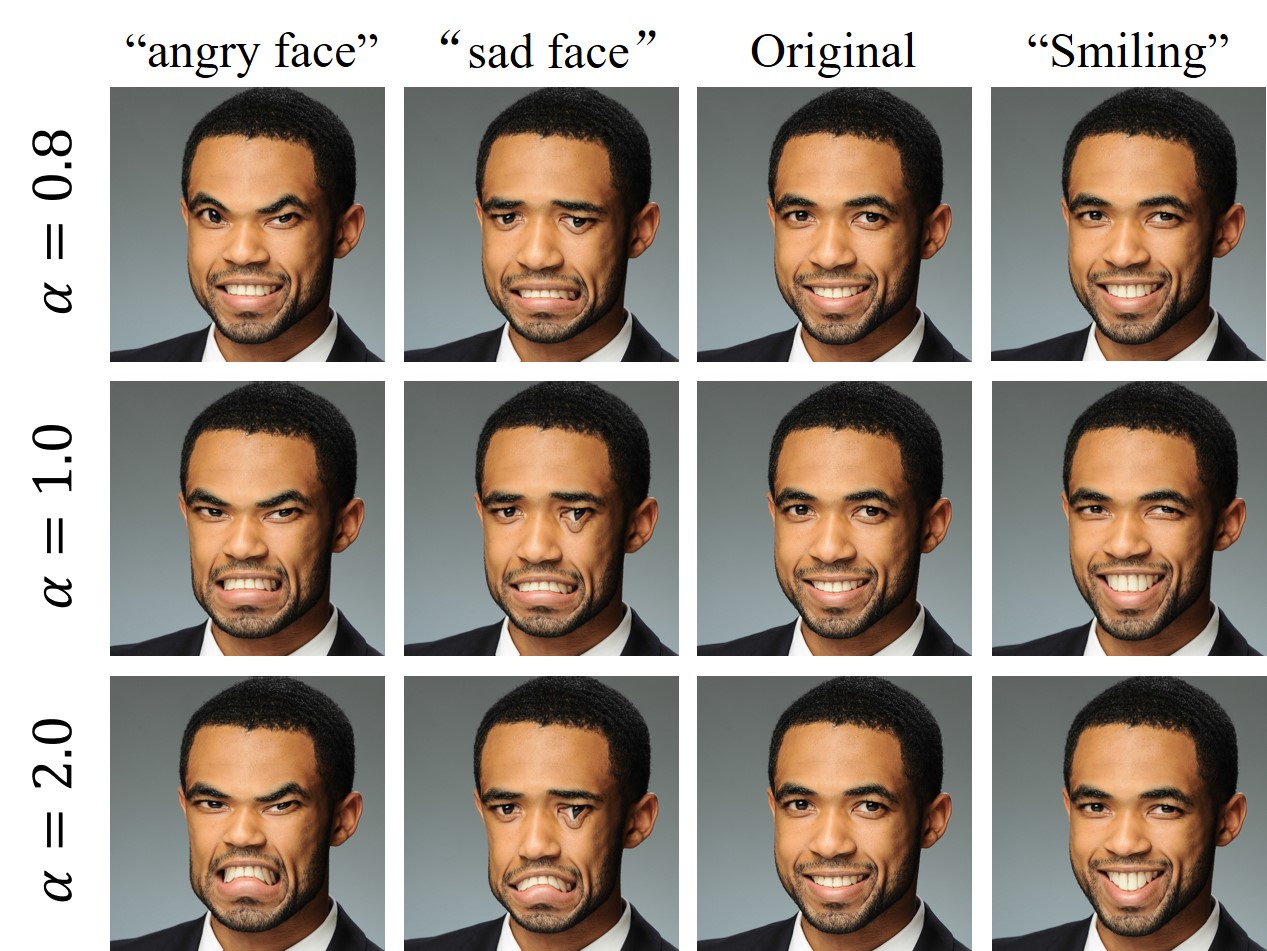}}\\
    \caption{\textbf{Amplitude of warping}. Different columns represent text prompts ranging from angry emotion to happy emotion, while the rows represent different warping amplitude.}
    \label{fig:amplitude_of_variation}
\end{figure}

\subsection{Amplitude of warping}
We set a controlled variable $\alpha$ on the vector flow field $U_s, U_c$ generation stage. This parameter is used to control the amplitude of warping each iteration, and Eq. (1) in the main text can be written as:
\begin{equation}
\begin{split}
    & I^\prime = T_{wp}(I, U_s, U_c), \\
    & U_s, U_c = \alpha \times G(I, t),
\end{split}
\end{equation}

By changing $\alpha$ and setting different degrees of text prompt description, we can get results in Fig.~\ref{fig:amplitude_of_variation}. Ranging from "angry", and "sad" to the final "happy" in text prompts, we can see the changing process of the emotions. In practice, users can change $\alpha$ to autonomously determine the desired degree of facial deformation.

\subsection{Comparison of our iterative optimization and one-shot learning}
We evaluate the time consumption of our iterative optimization mode and one-shot learning mode. We use a single GTX 1080 Ti Graphic Card to measure the data. Our evaluation results are in Tab.~\ref{tab:compare_oneshot}. The iterative optimization mode doesn't need any training stage but costs longer inference time than the one-shot learning mode. Our two vector flow fields representative methods: explicit representation and implicit representation have nearly the same inference time per iter, which is because the CLIP model consumes most of the computation in the backpropagation phase. However, according to Fig.~\ref{fig: flow_int_compare_loss}, implicit representation has a faster convergence speed than explicit representation. Users can set less total iter to decrease total inference time when using implicit representation.

\subsection{Limitations}

Since our model relies on the pre-trained CLIP model for a joint vision-language embedding, some text prompts which cannot be well aligned to the images by CLIP, may not be expected to yield a visual result that faithfully reflects the text prompts. 

Besides, since our method is based on image warping and color transformation, our method struggles to generate visual elements that are not present in the original image (eg, putting glasses on a face or generating hair for a bald head). On the one hand, this may be a limitation of our method compared to GAN-based methods. However, on the other hand, this may also be one of our advantages. Since our method adopts a shape-color decoupling design and has strong controllability, it can be well adapted to faces with different skin tones.

%------------------------------------------------------------------------
\section{Conclusion}

We propose a physically interpretable face editing method based on an arbitrary text prompt. We regard the face editing process as imposing a vector flow field on the ordinary image, which is then fitted by the implicit neural representation to carry out continuous, smooth, and resolution-independent warping. We optimize the implicit neural representation network weights under the supervision of the Contrastive Language-Image Pretraining~(CLIP) model. Furthermore, we achieve one-shot face editing based on arbitrary text prompts. Controlled experiments suggest the effectiveness of our designs and the superiority of our method over state-of-the-art methods.

{\small
\bibliographystyle{ieee_fullname}
\bibliography{egbib}
}

\end{document}